\documentclass[letterpaper]{article} % DO NOT CHANGE THIS
\usepackage{aaai24}  % DO NOT CHANGE THIS
\usepackage{times}  % DO NOT CHANGE THIS
\usepackage{helvet}  % DO NOT CHANGE THIS
\usepackage{courier}  % DO NOT CHANGE THIS
\usepackage[hyphens]{url}  % DO NOT CHANGE THIS
\usepackage{graphicx} % DO NOT CHANGE THIS
\usepackage{natbib}  % DO NOT CHANGE THIS AND DO NOT ADD ANY OPTIONS TO IT
\usepackage{caption} % DO NOT CHANGE THIS AND DO NOT ADD ANY OPTIONS TO IT
\usepackage{algorithm}
\usepackage{algorithmic}
\usepackage{amsmath}

\usepackage{newfloat}
\usepackage{listings}
\DeclareCaptionStyle{ruled}{labelfont=normalfont,labelsep=colon,strut=off} % DO NOT CHANGE THIS
\floatstyle{ruled}
\newfloat{listing}{tb}{lst}{}
\floatname{listing}{Listing}

\title{High-Fidelity 3D Head Avatars Reconstruction through Spatially-Varying Expression Conditioned Neural Radiance Field}
\author{
    Minghan Qin\equalcontrib,
    Yifan Liu\equalcontrib,
    Yuelang Xu,
    Xiaochen Zhao,
    Yebin Liu,
    Haoqian Wang
}
\affiliations{

    Tsinghua University
}

\usepackage{bibentry}

\begin{document}

\maketitle

\begin{abstract}

One crucial aspect of 3D head avatar reconstruction lies in the details of facial expressions. Although recent NeRF-based photo-realistic 3D head avatar methods achieve high-quality avatar rendering, they still encounter challenges retaining intricate facial expression details because they overlook the potential of specific expression variations at different spatial positions when conditioning the radiance field. Motivated by this observation, we introduce a novel Spatially-Varying Expression (SVE) conditioning. The SVE can be obtained by a simple MLP-based generation network, encompassing both spatial positional features and global expression information. Benefiting from rich and diverse information of the SVE at different positions, the proposed SVE-conditioned neural radiance field can deal with intricate facial expressions and achieve realistic rendering and geometry details of high-fidelity 3D head avatars. Additionally, to further elevate the geometric and rendering quality, we introduce a new coarse-to-fine training strategy, including a geometry initialization strategy at the coarse stage and an adaptive importance sampling strategy at the fine stage. Extensive experiments indicate that our method outperforms other state-of-the-art (SOTA) methods in rendering and geometry quality on mobile phone-collected and public datasets.

\end{abstract}

\section{Introduction}

Reconstructing controllable and realistic 3D head avatars is beneficial in many applications, such as VR/AR, games, and teleconferencing. The facial expression details are crucial in achieving realistic 3D head avatars. Current 3D head avatar reconstruction methods can generate controllable head avatars from monocular videos. However, achieving an accurate geometry of facial expressions and nuanced and individualized details remains a substantial challenge.

\begin{figure}[t]
  \centering
   \includegraphics[width=1.0\linewidth]{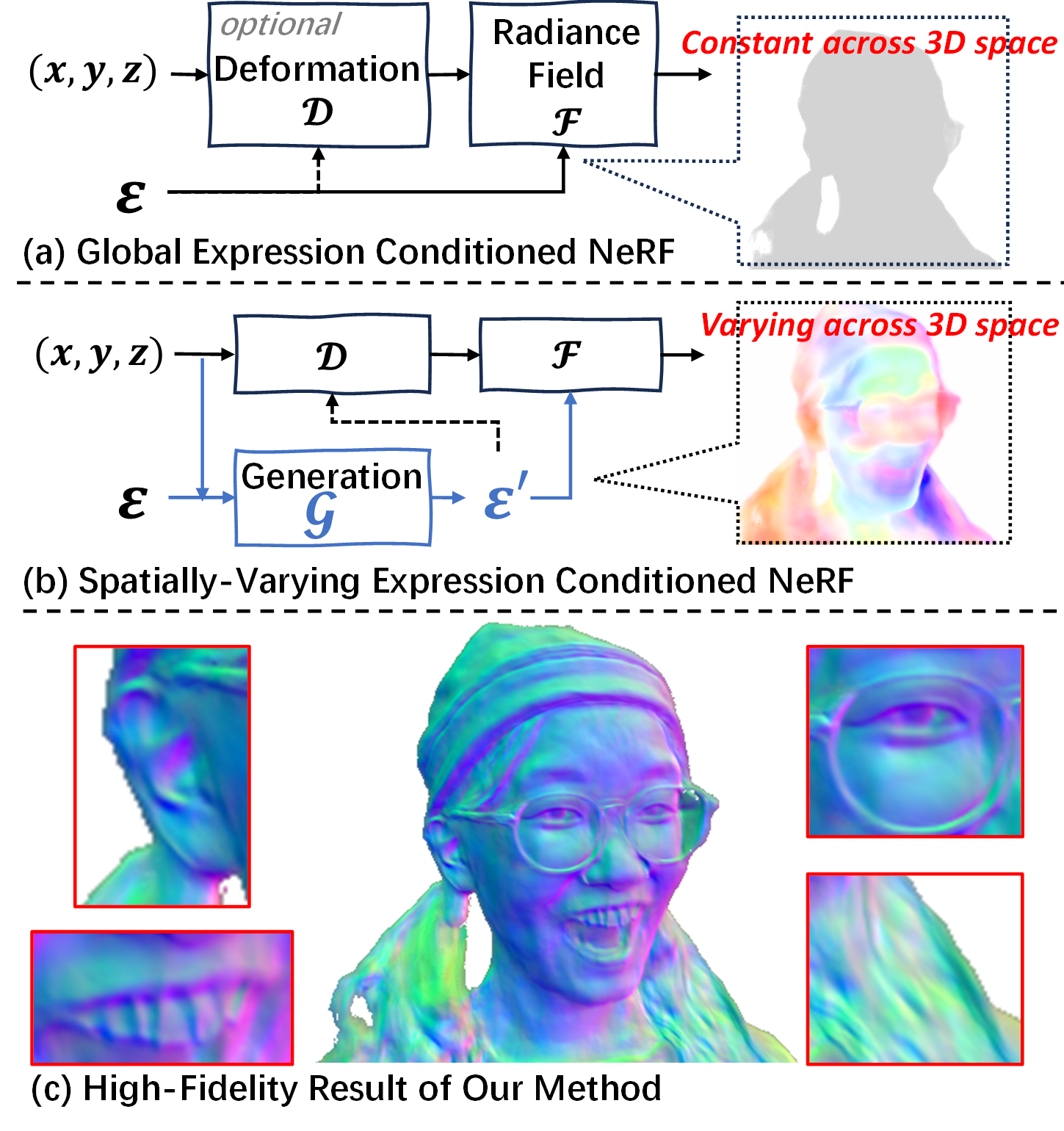}
   
   \caption{\textbf{Previous 3D head avatar methods based on (a) 3DMM global expression $\varepsilon$ conditioned NeRF (GE-NeRF) \textit{vs} (b)(c) Ours based on Spatially-Varying Expression $\varepsilon'$ conditioned NeRF (SVE-NeRF).} We visualize $\varepsilon$ and $\varepsilon'$ via volume rendering by replacing the RGB with $\varepsilon$ and $\varepsilon'$. The template expressions $\varepsilon$ used in GE-NeRF limit the $\mathcal{F}$ to capture expression and geometry details. While SVE-NeRF's $\varepsilon'$ incorporates expression and spatial positional features, guiding the $\mathcal{F}$ for enhanced expression rendering and geometry. We present high-quality geometric results of our method in (c).}
   \label{fig:intro}
\end{figure}

To reconstruct expressive 3D head avatars, some methods \cite{gafni2021dynamic, Gao2022nerfblendshape,athar2022rignerf,zheng2022avatar,INSTA:CVPR2023,xu2023avatarmav} based on neural radiance fields (NeRF) \cite{mildenhall2020nerf} achieves the photo-realistic rendering. However, these implicit neural radiance field-based approaches exhibit the \textbf{insufficient ability to render detailed complex expressions and the corresponding geometry}. As shown in Fig. \ref{fig:intro}, these methods typically employ an optional deformation network $\mathcal{D}$ to represent face expression motions and a NeRF $\mathcal{F}$ to model head geometry and appearance. Both $\mathcal{D}$ and $\mathcal{F}$ are conditioned on the \textbf{global expression} from 3DMM. Although recent methods \cite{athar2022rignerf,INSTA:CVPR2023} focus on better exploiting the global expression in an elaborately-designed deformation $\mathcal{D}$, such methods still directly utilize the global expression as the conditioning for NeRF $\mathcal{F}$. This direct global expression conditioning struggles to provide fine-grained control over the geometry and rendering at different positions within the 3D space. As a result, these methods struggles to obtain detailed rendering and accurate geometry when dealing with complex expressions.

To solve the above limitation, we propose \textbf{Spatially-Varying Expression} (SVE) as the conditioning. As indicated in Fig. \ref{fig:intro}, the global expression conditioning used in previous methods stays constant across 3D space. The global expression solely encompasses 3DMM template expressions, limiting $\mathcal{D}$ and $\mathcal{F}$ in capturing nuances and spatial intricacies, e.g., eyes, teeth, wrinkles (as shown in Fig. \ref{fig:comp1}). In contrast, SVE varies across 3D space, encompassing both spatial positional information and expression information. Therefore, SVE helps NeRF capture the intricate movement of wrinkles, eyes, mouth, eyebrows, etc. Specifically, to generate SVE, we design a simple generation network $\mathcal{G}$ to integrate the global expression parameters from the 3DMM with the spatial positional features of each position in 3D space. To reduce errors in geometry reconstruction from inadequate constraints and enhance overall quality, we introduce a coarse-to-fine training strategy to enhance the geometry at the coarse stage via the geometry initialization and improve the rendering quality at the fine stage by an adaptive importance sampling strategy. Extensive experiments demonstrate that our method achieves significantly superior results by employing the proposed SVE as conditioning in terms of accurate geometry reconstruction and detailed rendering, especially when dealing with intricate expressions. We present the contributions of our method as follows:

\begin{itemize}
    \item We propose a 3D head avatar method based on Spatially-Varying Expression conditioned neural radiance field. The proposed Spatially-Varying Expression (SVE) enables the radiance field to capture intricate expressions and detailed geometry faithfully.

    \item We design a simple generation network to generate the Spatially-Varying Expression by integrating the spatial positional features into the global expression. 
    
    \item We introduce a novel coarse-to-fine training strategy involving geometry initialization for credible reconstruction and adaptive importance sampling for enhanced rendering and geometry, thus refining avatar expression details.
   
\end{itemize}

\section{Related Work}

\textbf{3D Head Avatar Reconstruction.} Reconstructing 3D face models and head avatars has gained extensive research in recent years \cite{ichim2015dynamic, cao2015real, cao2016real, hu2017avatar, nagano2018pagan, athar2021flame, zheng2022avatar, athar2022rignerf, grassal2022neural, chan2022efficient, INSTA:CVPR2023, Zheng2023pointavatar, kirschstein2023nersemble, sun2023next3d}. Traditional 3DMM \cite{blanz1999morphable, gerig2018morphable} models appearance and geometry on linear space by PCA analysis. FLAME \cite{FLAME:SiggraphAsia2017} and other extension methods \cite{Feng:SIGGRAPH:2021, MICA:ECCV2022, EMOCA:CVPR:2021} additionally incorporate the eye and neck modelling, achieving the reconstruction of the whole head with vivid expressions and optimized texture.

As NeRF \cite{mildenhall2020nerf} shows great potential in photo-realistic rendering, some methods \cite{gafni2021dynamic, xu2023avatarmav,Gao2022nerfblendshape} explore NeRF to reconstruct controllable 3D head avatars by conditioning the NeRF with the global 3DMM tracked expression parameters. \cite{gafni2021dynamic} directly using the conditional neural radiance field without using any deformation networks. \cite{Gao2022nerfblendshape} utilize multiple multi-level hash tables to represent a specific expression. \cite{xu2023avatarmav} leverages a lightweight deformation network with voxel features generated by the global expression.

Some methods \cite{athar2022rignerf,zheng2022avatar,INSTA:CVPR2023} have also explored the combination of traditional 3DMM and neural rendering by leveraging the 3DMM face template as a prior for deformation. \cite{athar2022rignerf} and \cite{INSTA:CVPR2023} leverage the tracked face mesh templates to guide the deformation network. \cite{zheng2022avatar} proposes implicit mophorable models to incorporate 3DMM into volume rendering framework. These methods directly exploit the head geometry estimated by 3DMM, thus avoiding incorrect head geometry, such as facial concavities. However, since these methods rely on the 3DMM geometry prior, incorrect and excessively smooth surface estimation prevents the model from learning detailed, intricate expressions, hair, accessories, and clothing, leading to coarse geometry and rendering without rich details. 

Overall, the methods discussed above neglect leveraging the expressions variations at different spatial positions for modelling detailed geometry and appearances. In contrast, our methods explore the potential of the Spatially-Varying Expression conditioning, leading to detailed geometry and rendering results even with intricate expressions.

Recently researchers also have explored training a general generative human head model from large-scale datasets \cite{chan2022efficient, sun2023next3d, sun2022ide, wang2022morf, zhuang2022mofanerf, hong2022headnerf}, audio-driven talking head avatars \cite{guo2021ad,liu2022semantic}, avatars from dense multi-view data \cite{ma2021pixel,lombardi2018deep,lombardi2019neural,chu2020expressive,lombardi2021mixture,cao2022authentic}, and one-shot head avatars \cite{drobyshev2022megaportraits}, which is beyond the research topic of this work.

\textbf{Dynamic Neural Radiance Field.} We aim to reconstruct controllable 3D head avatars from monocular RGB videos. To model monocular RGB videos, we employ NeRF-based dynamic scene modelling methods \cite{Cao2022FWD, kplanes_2023,pumarola2021d,park2021nerfies,park2021hypernerf, fang2022fast} extend static NeRF \cite{mildenhall2020nerf} to model dynamic scenes by adding additional temporal information and deform fields. D-NeRF \cite{pumarola2021d} leverages the scene encoder to estimate the scene offsets between pre-defined canonical space and the current observation space from temporal embedding. Deformable NeRF \cite{park2021nerfies} explores a dense SE(3) deform field conditioned on frame-wise learnable latent codes. HyperNeRF \cite{park2021hypernerf} extends Deformable NeRF in terms of topological changes problem since the continuity of dense deform field can not model discontinuous topological changes. 

Recent methods \cite{Cao2022FWD, kplanes_2023} extend efficient triplane representation \cite{chan2022efficient} to dynamic scenes and model dynamic scenes without explicit deform fields for accelerating. Unlike existing dynamic NeRFs, we design the deformation as a tiny network, directly utilizing the Spatially-Varying Expression to predict 6D motions. Benefiting from the rich information of Spatially-Varying Expression, our method achieve competitive results compared to methods with well-signed deform network. In addition to the former dynamic scene modelling methods, our method can control various expressions for self and cross-identity reenactment.

\section{Preliminary}
Our method is based on the neural radiance field (NeRF), combining the conditional NeRF and the deformable NeRF for better avatar motion control.

\textbf{NeRF} \cite{mildenhall2020nerf} is an implicit function-based volumetric rendering technique which enables photo-realistic novel view synthesis. The radiance field $\mathcal{F}$ maps 3D spatial query points $p=(x, y, z)$ and the corresponding 2D view direction $d$ to density $\sigma$ and color $c$.

\begin{equation}
    \sigma, c = \mathcal{F}_{\theta_{F}}(p, d)
\end{equation}
where $\theta_{F}$ is learnable parameters of the radiance field $\mathcal{F}$. By computing the density $\sigma$ and color $c$ of each query point $p$ along a ray from the camera origin $o$ through a pixel $p_{2d}=(u, v)$, the RGB value of the pixel $p$ can be obtained through integration of volume rendering.

\textbf{Conditional NeRF} \cite{gafni2021dynamic} with additional parameters $\varepsilon$ enables NeRF's adaptability to varying condition information. In 3D head avatars reconstruction, most methods utilize pre-tracked global 3DMM expression parameters $\varepsilon$ as the conditioning. The condition process is usually implemented by direct concatenation or addition of encoded $p$ and $\varepsilon$.

\begin{equation}
\label{eq:condition}
    \sigma, c = \mathcal{F}_{\theta_{{F}}}(p, d, \varepsilon)
\end{equation}

\textbf{Deformable NeRF} \cite{park2021nerfies} disentangles the shape and motion for dynamic scenes. Deformable NeRF introduces a dense deformation field $\mathcal{D}$ to deform query points $p_o=(x, y, z)$ from the observation space of the current frame to a canonical space $p_c=(x', y', z')$, then estimate the radiance field $\mathcal{F}$ in the canonical space.

\begin{equation}
    \begin{aligned}
    p_c &= \mathcal{D}_{\theta_{{D}}}(p_o, t) \\
    \sigma, c &= \mathcal{F}_{\theta_{{F}}}(p_c, d)
    \end{aligned}
\end{equation} 
where $\theta_{{D}}$ denotes learnable parameters of the dense deformation field $\mathcal{D}$. $t$ represents a certain frame of the dynamic scene.

\begin{figure*}[t]
  \centering

   \includegraphics[width=0.9\linewidth]{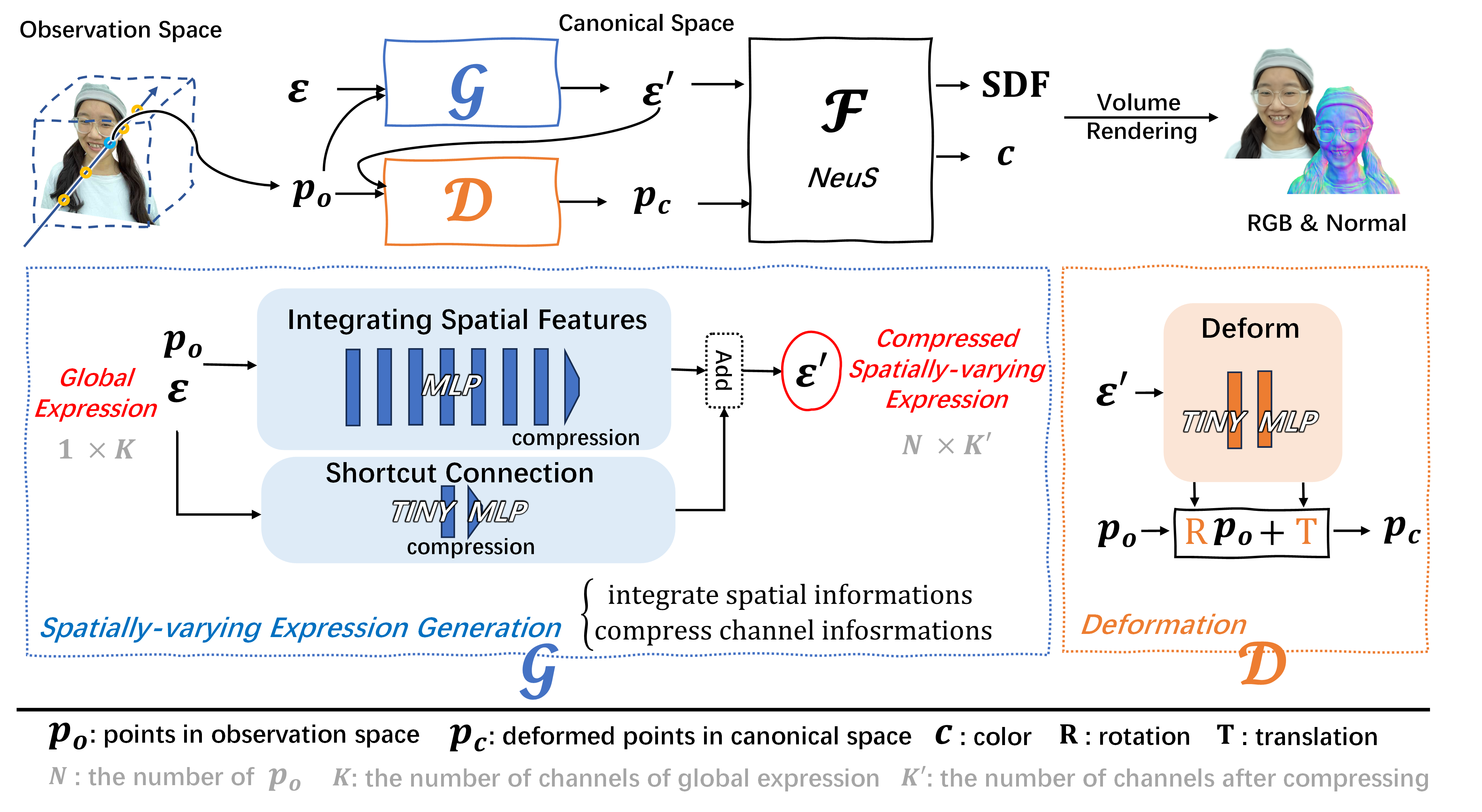}

   \caption{
   \textbf{Method Overview.} Given a portrait video, We first track the global expression parameters $\varepsilon$ using 3DMM \cite{gerig2018morphable}. After the pre-processing, given the sampled 3D points $p_o$ in observation space, we apply the generation network $\mathcal{G}$ to extend the global expression parameters $\varepsilon$ with the spatial positional features of each position $p_o$ in 3D space. Then, through a deformation network $D$, we transform $p_o$ from the observation space to the $p_c$ in the canonical space conditioned on $\varepsilon'$. Subsequently, we use $\varepsilon'$ conditioned NeuS \cite{wang2021neus} to predict the $\text{SDF}$ values and color $c$ corresponding to $p_c$. Finally, we obtain the rendered RGB image and normal using volumetric rendering.}
   \label{fig:overview}
\end{figure*}

\section{Method}

\subsection{Formulation}
\label{sec:formulation}

Previous methods have exploited the power of NeRF for photo-realistic 3D head avatar reconstruction. A typical NeRF-based head avatar model can be formulated as an expression-conditioned deformable NeRF as Eq. \ref{eq:formulation1}.

\begin{equation}
\label{eq:formulation1}
    \begin{aligned}
    p_c &= \mathcal{D}_{\theta_{{D}}}(p_o| \varepsilon) \\
    \sigma, c &= \mathcal{F}_{\theta_{{F}}}(p_c, d | \varepsilon)
    \end{aligned}
\end{equation}
where $\varepsilon$ denotes the global expression parameters obtained by tracking the input video using 3DMM face templates. Previous methods condition the radiance field $\mathcal{F}$ directly on the global expression parameters $\varepsilon$, neglecting the expression conditioning for specific spatial positions. Therefore, this per-frame global expression conditioning is insufficient to obtain detailed expression rendering and geometry. 

Motivated by this observation, we present a novel approach utilizing the Spatially-Varying Expression (SVE) conditioned NeRF. Our method can be formulated as Eq. \ref{eq:formulation2}.

\begin{equation}
\label{eq:formulation2}
    \begin{aligned}
    \varepsilon' &= \mathcal{G}_{\theta{{G}}}(\varepsilon | p_o) \\
    p_c &= \mathcal{D}_{\theta_{{D}}}(p_o| \varepsilon') \\
    \text{SDF}, c &= \mathcal{F}_{\theta_{{F}}}(p_c, d | \varepsilon')
    \end{aligned}
\end{equation}
where $p_o=(x, y, z)$ is the query points in the observation space. $p_c=(x', y', z')$ is the query points deformed by network $\mathcal{D}$ in the canonical space. We select NeuS \cite{wang2021neus} as $\mathcal{F}$ , which incorporates Signed Distance Function (SDF) as the implicit representation. And $c$ is the predicted color of each query point. $C = f(A | B)$ means the function $f$ maps $A$ to $C$ conditioned on $B$. In contrast to global expression parameters $\varepsilon$, $\varepsilon'$ denotes the generated compressed Spatially-Varying Expression parameters via a simple generation network $\mathcal{G}$. Through the network $\mathcal{G}$, we effectively integrate a certain frame's global expression with the spatial positional information, acquiring the Spatially-Varying Expression. 

\subsection{Spatially-Varying Expression Conditioned NeRF}
\label{sec:generation}

\textbf{Overview.} As depicted in Fig. \ref{fig:overview}, given a certain training frame, we first extract the per-frame global expression parameters $\varepsilon$ by 3DMM  pre-processing. During the training, We first obtain rays according to camera poses and head poses. Then, after sampling a set of points $p_o$ in the current frame observation space, we utilize the proposed generation network $\mathcal{G}$ to generate the Spatially-Varying Expression (SVE) parameters $\varepsilon'$ with $p_o$ as the additional condition to provide spatial positional features. Subsequently, we leverage a tiny deformation network $\mathcal{D}$ to specify the 6D motion $\text{R}, \text{T}$ of each query point $p_o$ according to the generated SVE parameters $\varepsilon'$, and deform $p_o$ to the query points in canonical space $p_c$. The deformation module $\mathcal{D}$ is designed as a remarkably simple structure without leading to performance degradation due to benefiting from the spatial information contained in $\varepsilon'$. Subsequently, the neural radiance field $\mathcal{F}$ based on NeuS takes the deformed query points $p_c$ as inputs and the SVE parameters $\varepsilon'$ as the conditioning, yielding the SDF and color values $c$. Finally, by employing the volumetric rendering technique, we integrate the color values $c$ of each ray to obtain the rendered RGB of each pixel. In the meanwhile, we also integrate the gradient of each point's SDF value to get the rendered normal map.

\textbf{Spatially-Varying Expression generation.} The proposed term \textbf{Spatially-Varying} refers to changing across different spatial positions. Previous methods \cite{athar2022rignerf, INSTA:CVPR2023} directly utilize the global expression parameters from 3DMM when modelling avatars' geometry and appearance. These methods focus on enhancing the utilization of the global expression parameters during the deformation $\mathcal{D}$, neglecting the potential to fully leverage the information contained in expressions for modelling both geometry and appearance. However, various facial expressions influence the intricacies of facial geometry and texture. For instance, a laughing expression does not only involve the mouth being open. The muscles' motion throughout the face, the raised angle of eyebrows and skin-creasing should exhibit varying levels of geometric and textural changes. Even with well-designed deformation networks, previous methods are challenging to capture such nuanced expression changes. This is because they overlook the potential of injecting spatial information into the global expression conditioning of the radiance field. Consequently, these prior methods based on the global expression inhibit the capacity of the neural radiance field to learn each 3D position's distinct features of the expression. 

In contrast, our proposed expression parameters explore the potential of the radiance field to learn distinct features for different positions across the 3D space. To elaborate, the deformations ($R$ and $T$) and characteristics ($\text{SDF}$ and $c$) associated with each point $p=(x,y,z)$ in 3D space should rely not only on its spatial position $(x, y, z)$ but also on the global expression parameters $\varepsilon$. Considering that this variable changes in conjunction with alterations in 3D positions, it is referred to as \textbf{Spatially-Varying Expression}.

Specifically, as shown in Fig. \ref{fig:overview}, the generation module $\mathcal{G}$ comprises two networks. The spatial feature integrating network amalgamates the spatial positional information contained in query points $p_o$ with the expression parameters. This network learns the influence of the global expression parameters on these specific positions $p_o=(x,y,z)$. Simultaneously, the shortcut connection compresses the global expression parameters $\varepsilon$ for residual addition. While the integrating network is parameterized as a Multilayer Perceptron (MLP) consisting of 8 fully-connected layers, the shortcut connection only consists of 2 fully-connected layers, serving as dimensional mapping for residual addition. Through the generation module $\mathcal{G}$, we obtain the Spatially-Varying Expression parameters, which encapsulate both the global expression information and the spatial positional features.

Note that in both the integrating network and the shortcut connection, we reduce the dimension of the outputs from $K$ to $K'$ as shown in Fig. \ref{fig:overview} to avoid over-fitting. Intuitively, one high-dimensional global expression parameter can sufficiently capture the coarse changes of facial expressions validated by previous methods discussed above. However, when employing Spatially-Varying Expression as conditioning for $D$ and $F$, it degrades performance when generalizing to new expressions due to the excessive information for each query points $p_o$. Therefore, by reducing the dimension of the Spatially-Varying Expression codes, we effectively avoid over-fitting without performance degradation.

\textbf{Expression conditioned deformation.} As we discussed above, the 6D motion deformation $\text{R}, \text{T}$ of each point $p_o=(x,y,z)$ in the 3D observation space should rely on the generated Spatially-Varying Expression $\varepsilon'$. Therefore, we design the deformation as a lightweight tiny MLP consisting of two fully-connected layers. Compared to the well-designed deformation modules proposed by prior methods, our deformation network $\mathcal{D}$ benefits from the rich position-dependent expression information from the Spatially-Varying Expression parameters, thus effectively predicting accurate 6D motions with only a lightweight network.

\begin{figure*}[t]
  \centering

   \includegraphics[width=1.0\textwidth]{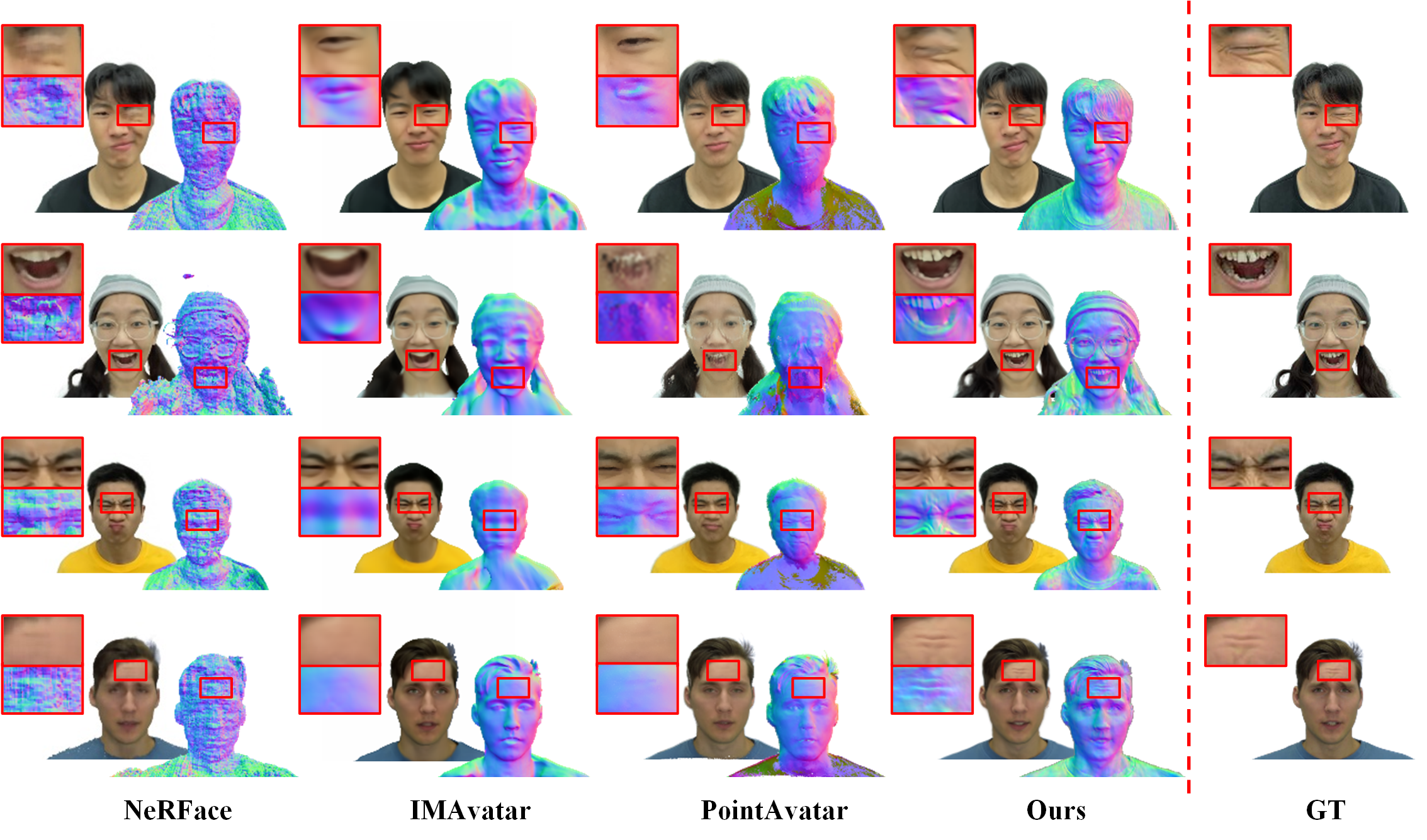}
   \caption{\textbf{Qualitative comparisons on self reenactment task.} From left to right: NeRFace \cite{gafni2021dynamic}, IMAvatar \cite{zheng2022avatar}, PointAvatar \cite{Zheng2023pointavatar}, and Ours. Our method reconstructs high-quality rendering and geometric details of wrinkles, teeth, hairs and accessories. \textbf{We recommend zooming in to see more details.}}
   \label{fig:comp1}

\end{figure*}

\subsection{Coarse-to-Fine Training Strategy}
\label{sec:strategy}

\textbf{Geometry initialization of coarse stage.} In monocular 3D head avatar reconstruction, the geometry often encounters geometric collapse: the facial structure deviations from the intended geometry, leading to varying degrees of concavity. Previous methods \cite{zheng2022avatar, Zheng2023pointavatar} try to address this issue by incorporating 3DMM templates as a prior into the deformation network $\mathcal{D}$ to constrain the geometry. Nonetheless, the inherent smoothness and limitation of the 3DMM itself lead to lacking intricate geometry details of facial features, hair, clothing, etc. 

To tackle this issue, we present an innovative geometry initialization strategy at the coarse training stage to achieve a harmonious equilibrium between intricacies and smoothness geometries. We harness the pseudo-depth of tracked 3DMM model to safeguard the geometry against geometric collapse. Instead of using strong 3DMM constraints throughout the training, the initialization strategy avoids excessively smooth geometry.

Specifically, in the geometry initialization, we employ the rendered pseudo-depth $D$ of the tracked 3DMM template, as shown in the supplementary appendix. We map the sampled pixels of rendered pseudo-depth into the 3D observation space points $p_{o}^d$. Then, we predict the SDF values $\text{SDF}^d$ corresponding to $p_{o}^d$ according to Eq. \ref{eq:formulation2}. Because $p_{o}^d$ lie on the approximate surface of the face, their associated SDF values $\text{SDF}^d$ should naturally tend towards zero. To enforce this, we employ a geometry loss to constrain the predicted $\text{SDF}^d$. Furthermore, by utilizing an L1 loss, we also guide the alignment of the rendered depth $\hat{D}$ with its corresponding pseudo-depth $D$.

\textbf{Adaptive importance sampling strategy of the fine stage.} The adaptive importance sampling strategy is proposed to achieve a sensitive perception of infrequent areas and small areas, e.g., teeth and rim glasses. In contrast to commonly used random pixel sampling \cite{mildenhall2020nerf}, importance sampling with fixed-weight \cite{gafni2021dynamic} or with the pre-computed weight \cite{li2022neural}, our strategy automatically adapts to different training data, and dynamically adjust the weight during the training.

Specifically, for each training frame, we first segment the frames into $N=19$ semantic regions, encompassing various components such as eyes, face, hair, lips, etc. Specific classification criteria are detailed in the supplementary materials. We assign weights ${w_i^s}, i=1,2,\dots, N$ to each region at the $s$-th training step. During the $s$-th training step, we calculate the guidance loss ${L_i^s}$ for sampled points within each region based on Eq. \ref{eq:guide-loss} and the area of each region ${A_i^s}$ of the current frame.

\begin{equation}
    \begin{aligned}
        L_i^s &= \lambda_1 L_{i_{\text{,render}}}^s + \lambda_2 L_{i_{,\text{depth}}}^s \\
        L_{i_{,\text{render}}}^s &= M_i\|\hat{C}_i - C_i\|_1 + BCE(\hat{M_i}, M_i) \\
        L_{i_{,\text{depth}}}^s &= M_i\|\hat{D}_i - D_i\|_1
    \end{aligned}
    \label{eq:guide-loss} 
\end{equation} 
Here, $M_i$, $C_i$, and $D_i$ stand for the ground-truth mask, color, and pseudo-depth of the region $i$. $\hat{M}$, $\hat{C}$, and $\hat{D}$ represent the corresponding predictions. The values $\lambda_1$ and $\lambda_2$ help decide whether $L_i^s$ should concentrate more on parts with unsatisfactory rendering or areas with subpar geometry.

Next, we utilize the loss $L_i^s$ for guidance to update ${w_i^s}$ using exponential moving average (EMA) according to \ref{eq:weight-update}. The EMA updating stabilises the updating and also addresses the issue of not being able to resample region $i$ in a training step when its area $A_i^{s}$ becomes 0. Because when $A_i^{s}$ is 0, the updated weight $w_i^{s+1}$ also becomes 0 without EMA. Consequently, in subsequent training steps, both $L_i^{s+m}$ and $w_i^{s+m}$ with $m \geq 1$ remain 0. In such cases, the importance sampling will not sample points within region $i$. 

\begin{equation}
\label{eq:weight-update}
w_i^{s+1}=\left(\frac{L_i^{s}}{ w_i^{s} A_i^{s}\sum_i L_i^{s}}\right) \cdot \alpha+ w_i^{s} \cdot(1-\alpha)
\end{equation}
where $\alpha$ is the updating ratio, set to 0.01 empirically. This loss-guided sampling strategy adaptively encourages the model to prioritize previously inadequately learned regions, leading to improved expression realism and rendering quality.

\section{Experiments}
\label{sec:exp}

\subsection{Datasets and Preprocessing}

\textbf{Datasets.} We collect seven monocular RGB sequences of different subjects. All videos are collected using an iPhone 12 front camera with a fixed camera pose and a length of 2000-4000 frames. The image resolution of each video is 480$\times$480. The content of each collected video includes facial expression changes, head pose changes, and talking. Additionally, to evaluate the effectiveness of our method on public datasets, we also conduct experiments on two open-source datasets from IMAvatar \cite{zheng2022avatar} and NeRFace \cite{gafni2021dynamic}.

\textbf{Pre-processing.}
During the pre-processing, we exploit PP-Matting \cite{chen2022pp} to generate foreground masks and the face parsing method \cite{faceparse} to obtain coarse semantic segmentation maps for subsequent adaptive importance sampling. The BFM Model \cite{gerig2018morphable} is used to track the collected video's head poses and expression parameters. We render the predicted pseudo-depth from BFM-tracked face meshes. 

\begin{figure}[t]
  \centering
   \includegraphics[width=1.0\linewidth]{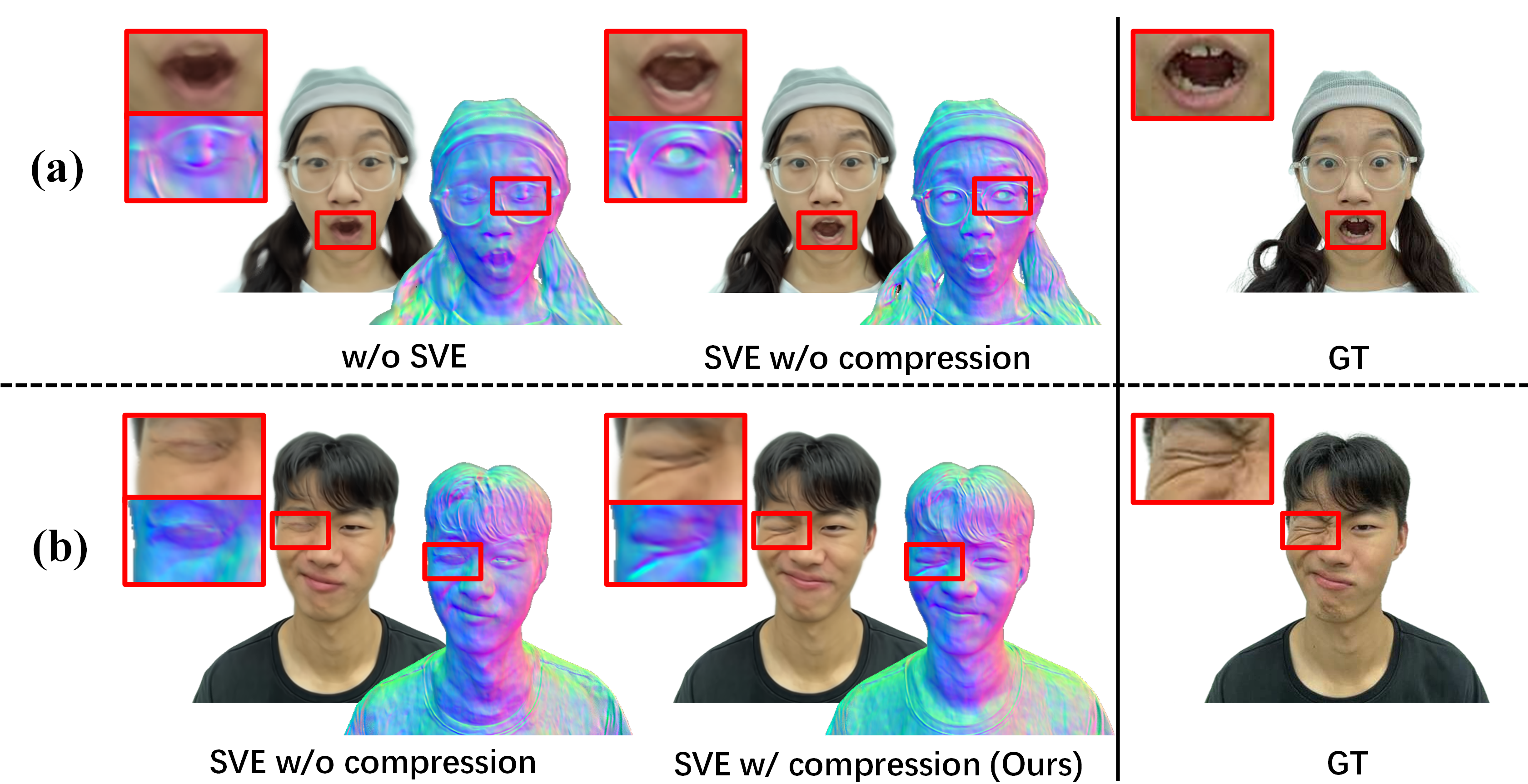}
   \caption{\textbf{Qualitative ablation results of Spatially-Varying Expression (SVE).} }
   \label{fig:ablation123}
\end{figure}

\begin{figure}[t]
  \centering
   \includegraphics[width=1.0\linewidth]{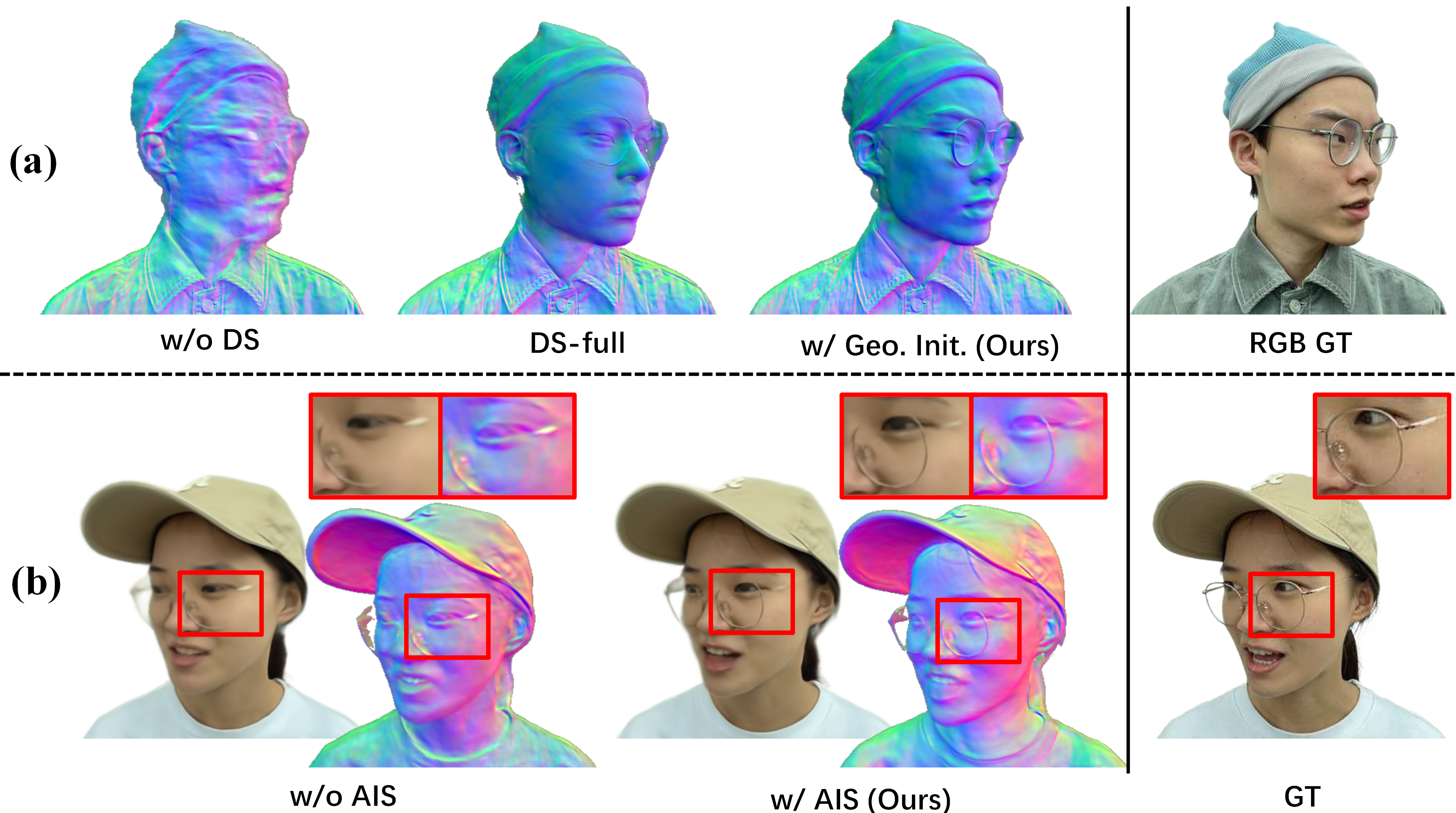}
   \caption{\textbf{Qualitative ablation results of the geometry initialization (Geo. Init.) strategy and the adaptive importance sampling (AIS).} In the case w/o DS, the concavity and convexity of the face are incorrect.}
   \label{fig:ablation45}
\end{figure}

\subsection{Comparison on Avatar Reconstruction Quality}
We conducted qualitative and quantitative comparisons with three SOTA NeRF-based 3D head avatar reconstruction methods: NeRFace \cite{gafni2021dynamic}, IMAvatar \cite{zheng2022avatar}, and PointAvatar \cite{Zheng2023pointavatar}. NeRFace directly uses global expression conditioned NeRF without explicit deformation. IMAvatar employs implicit morphing based on the FLAME head templates \cite{athar2021flame} in the deformation network to utilize the expression parameters. PointAvatar extends IMAvatar by utilizing a point-based neural representation approach for efficient training. We recommend reading the supplementary appendix and video for more experimental results, including novel view synthesis, cross-identity reenactment, and additional comparison results.

The qualitative results of the self-reenactment task are shown in Fig. \ref{fig:comp1}. Results suggest that IMAvatar and PointAvatar reconstruct coarse head geometry. Constrained by the smooth 3DMM template, these two approaches struggle to capture intricate expression details, especially on our datasets with complex expression variations. NeRFace relies on global expression parameters, limiting its incorporation of distinct expression information across spatial positions. Consequently, it struggles to render accurate, fine-grained expression details. Moreover, without reliance on prior 3DMM template knowledge and geometric optimization, NeRFace fails to accurately reconstruct facial geometry. In contrast, our method leverages Spatially-Varying Expression as the conditioning of the radiance field, effectively harnessing the information embedded within 3DMM expression parameters and spatial positional features. As a result, our method markedly outperforms the previous SOTA methods in terms of expression detail refinement and geometric reconstruction quality.

As for quantitative comparisons, We evaluate the rendering quality and expression similarity of all the methods discussed above. Tab. \ref{table_comp} reports several commonly used metrics for rendering quality evaluation, including Mean Absolute Error (MAE), Peak Signal-to-Noise Ratio (PSNR), Structure Similarity Index (SSIM), and Learned Perceptual Image Patch Similarity (LPIPS) \cite{zhang2018unreasonable}. We computed the average performance across nine subjects. Our method notably outperforms all SOTA approaches. Furthermore, in the supplementary appendix, we provide individual quantitative comparison results for all subjects to offer more compelling evidence for the comparisons.

\begin{table}[t]
\centering
\caption{\textbf{Quantitative evaluations on self-reenactment task.} We show the Average performance on all nine subjects. Our method notably outperforms all SOTA approaches. For separate results of individual subjects, please refer to the supplementary appendix.}
\resizebox{1.0\linewidth}{!}{
\begin{tabular}{ccccc}
\hline
            & L1↓               & PSNR↑             & SSIM↑             & LPIPS↓            \\ \hline
NeRFace     & 0.0195          & 24.976          & 0.933          & 0.122          \\
IMavatar    & 0.0187           & 25.360          & 0.927          & 0.145          \\
PointAvatar & 0.0207          & 24.799          & 0.918          & 0.130          \\
Ours        & \textbf{0.0148} & \textbf{27.748} & \textbf{0.944} & \textbf{0.0925} \\ \hline
\end{tabular}
}
\label{table_comp}
\end{table}

\subsection{Ablation Study}
\label{sec:ablation}

\textbf{Effectiveness of the Spatially-Varying Expression (SVE).} To validate the effectiveness of SVE and the compression of SVE, we show the comparisons between our methods and these two baselines: (1) \textbf{w/o SVE.} We use the tracked 64-dimensional 3DMM expression parameters as the NeRF's condition. (2) \textbf{SVE w/o compression.} We replace the shortcut compression in the generation network $\mathcal{G}$ to an identity mapping and modify the integrating network of $\mathcal{G}$ to retain 64-dimensional features. Then we consider the addition of the output of the two branches as the condition. Please refer to the supplementary appendix for detailed illustrations of these two baselines. As depicted in Fig. \ref{fig:ablation123}(a) demonstrates that employing Spatially-Varying Expression (SVE) without compression can exploit spatial positional information, resulting in clearer reconstructed expression details than using global expression only. As depicted in Fig. \ref{fig:ablation123}(b) demonstrates, compression of SVE mitigates the risk of over-fitting and improves expression detail reconstruction quality.

\textbf{Effectiveness of the coarse-to-fine training strategy.}To validate the effectiveness of the geometry initialization strategy and the adaptive importance sampling (AIS) in coarse-to-fine training strategy, we show the comparisons between our methods and these three baselines: (1) \textbf{w/o Depth Supervision (w/o DS).} We refrain from employing the predicted depth as supervision. (2) \textbf{Depth Supervision during full training stage (DS-full).} We incorporate the predicted depth as supervision throughout the training process. (3) \textbf{w/o AIS.} We employ a random sampling approach to sample pixels in training. As depicted in Fig. \ref{fig:ablation45}(a), the approach employing geometry initialization accomplishes refined geometric reconstruction. As depicted in Fig. \ref{fig:ablation45}(b), the adaptive importance sampling directs the network's focus towards overlooked intricate regions.

\begin{table}[t]
\centering
\caption{\textbf{Quantitative results of ablation studies.} Our method achieves better performance compared to the baselines.}
\resizebox{1.0\linewidth}{!}{
\begin{tabular}{cccccc}
\hline   & L1↓       & PSNR↑    & SSIM↑ & LPIPS↓ \\ \hline
w/o SVE & 0.0198   & 24.098   & 0.922 & 0.0963 \\
SVE w/o compress & 0.0188   & 24.090   & 0.925 & 0.0923 \\
w/o DS  & 0.0199 & 23.707 & 0.920 & 0.0933 \\
DS-full  & 0.0190 & 24.069 & 0.924 & 0.0925 \\
w/o AIS  & 0.0191 & 24.011   & 0.921 & 0.0931 \\
ours  & \textbf{0.0187} & \textbf{24.335} & \textbf{0.927} & \textbf{0.0911} \\ \hline
\end{tabular}
}
\label{table_ablation}
\end{table}

\section{Conclusions}
In this paper, we have proposed a 3D head avatars reconstruction method through Spatially-Varying Expression conditioned NeRF. The Spatially-Varying Expression (SVE) integrates global expression with localized spatial positional features, enabling the radiance field to capture intricate expressions and geometric details accurately. We employ a concise yet effective MLP-based generation network to produce the compressed SVE  by integrating spatial positional features with the global expression from 3DMM. Furthermore, we introduce an innovative coarse-to-fine training strategy, including a geometry initialization technique and adaptive importance sampling strategy, thus further refining the expression details of avatars. Prior to this work, there has been a lack of focus on efficiently leveraging the global expression to achieve improved conditioned NeRF for reconstruction quality. We aspire for this study to garner attention from researchers and instigate ongoing explorations in this direction.

\section{Discussion}

\textbf{Limitation.} Despite achieving high-quality reconstruction of 3D head avatars, the generalization capacity of our method remains constrained by the distribution of data. Our method encounters challenges in generating distinct teeth when the dataset predominantly comprises instances of mouth opening (with a video length proportion below 5\%). We will show our failure cases in the supplementary appendix.

\textbf{Future Work.} To address the above shortcomings, we will try to train a 3D head avatar reconstruction model with enhanced expression generalization capabilities using large-scale facial video datasets.

\bibliography{aaai24}

\begin{thebibliography}{48}
\providecommand{\natexlab}[1]{#1}

\bibitem[{Athar, Shu, and Samaras(2023)}]{athar2021flame}
Athar, S.; Shu, Z.; and Samaras, D. 2023.
\newblock Flame-in-nerf: Neural control of radiance fields for free view face
  animation.
\newblock In \emph{IEEE 17th International Conference on Automatic Face and
  Gesture Recognition (FG)}, 1--8.

\bibitem[{Athar et~al.(2022)Athar, Xu, Sunkavalli, Shechtman, and
  Shu}]{athar2022rignerf}
Athar, S.; Xu, Z.; Sunkavalli, K.; Shechtman, E.; and Shu, Z. 2022.
\newblock Rignerf: Fully controllable neural 3d portraits.
\newblock In \emph{Proceedings of the IEEE/CVF Conference on Computer Vision
  and Pattern Recognition}, 20364--20373.

\bibitem[{Blanz and Vetter(1999)}]{blanz1999morphable}
Blanz, V.; and Vetter, T. 1999.
\newblock A morphable model for the synthesis of 3D faces.
\newblock In \emph{Proceedings of the 26th annual conference on Computer
  graphics and interactive techniques}, 187--194.

\bibitem[{Cao and Johnson(2023)}]{Cao2022FWD}
Cao, A.; and Johnson, J. 2023.
\newblock HexPlane: A Fast Representation for Dynamic Scenes.
\newblock \emph{CVPR}.

\bibitem[{Cao et~al.(2015)Cao, Bradley, Zhou, and Beeler}]{cao2015real}
Cao, C.; Bradley, D.; Zhou, K.; and Beeler, T. 2015.
\newblock Real-Time High-Fidelity Facial Performance Capture.
\newblock \emph{ACM Trans. Graph.}, 34(4).

\bibitem[{Cao et~al.(2022)Cao, Simon, Kim, Schwartz, Zollhoefer, Saito,
  Lombardi, Wei, Belko, Yu, Sheikh, and Saragih}]{cao2022authentic}
Cao, C.; Simon, T.; Kim, J.~K.; Schwartz, G.; Zollhoefer, M.; Saito, S.-S.;
  Lombardi, S.; Wei, S.-E.; Belko, D.; Yu, S.-I.; Sheikh, Y.; and Saragih, J.
  2022.
\newblock Authentic Volumetric Avatars from a Phone Scan.
\newblock \emph{ACM Trans. Graph.}, 41(4).

\bibitem[{Cao et~al.(2016)Cao, Wu, Weng, Shao, and Zhou}]{cao2016real}
Cao, C.; Wu, H.; Weng, Y.; Shao, T.; and Zhou, K. 2016.
\newblock Real-Time Facial Animation with Image-Based Dynamic Avatars.
\newblock \emph{ACM Trans. Graph.}, 35(4).

\bibitem[{Chan et~al.(2022)Chan, Lin, Chan, Nagano, Pan, Mello, Gallo, Guibas,
  Tremblay, Khamis, Karras, and Wetzstein}]{chan2022efficient}
Chan, E.~R.; Lin, C.~Z.; Chan, M.~A.; Nagano, K.; Pan, B.; Mello, S.~D.; Gallo,
  O.; Guibas, L.; Tremblay, J.; Khamis, S.; Karras, T.; and Wetzstein, G. 2022.
\newblock Efficient Geometry-aware {3D} Generative Adversarial Networks.
\newblock In \emph{Proceedings of the IEEE/CVF Conference on Computer Vision
  and Pattern Recognition (CVPR)}, 16102--16112.

\bibitem[{Chen et~al.(2022)Chen, Liu, Wang, Peng, Hao, Chu, Tang, Wu, Chen, Yu
  et~al.}]{chen2022pp}
Chen, G.; Liu, Y.; Wang, J.; Peng, J.; Hao, Y.; Chu, L.; Tang, S.; Wu, Z.;
  Chen, Z.; Yu, Z.; et~al. 2022.
\newblock PP-Matting: High-Accuracy Natural Image Matting.
\newblock \emph{arXiv preprint arXiv:2204.09433}.

\bibitem[{Chu et~al.(2020)Chu, Ma, Torre, Fidler, and
  Sheikh}]{chu2020expressive}
Chu, H.; Ma, S.; Torre, F.; Fidler, S.; and Sheikh, Y. 2020.
\newblock Expressive Telepresence via Modular Codec Avatars.
\newblock In \emph{Proceedings of the Proceedings of the European Conference on
  Computer Vision (ECCV)}, 330--345.

\bibitem[{Danecek, Black, and Bolkart(2022)}]{EMOCA:CVPR:2021}
Danecek, R.; Black, M.~J.; and Bolkart, T. 2022.
\newblock {EMOCA}: {E}motion Driven Monocular Face Capture and Animation.
\newblock In \emph{Conference on Computer Vision and Pattern Recognition
  (CVPR)}, 20311--20322.

\bibitem[{Drobyshev et~al.(2022)Drobyshev, Chelishev, Khakhulin, Ivakhnenko,
  Lempitsky, and Zakharov}]{drobyshev2022megaportraits}
Drobyshev, N.; Chelishev, J.; Khakhulin, T.; Ivakhnenko, A.; Lempitsky, V.; and
  Zakharov, E. 2022.
\newblock MegaPortraits: One-shot Megapixel Neural Head Avatars.

\bibitem[{Fang et~al.(2022)Fang, Yi, Wang, Xie, Zhang, Liu, Nie\ss{}ner, and
  Tian}]{fang2022fast}
Fang, J.; Yi, T.; Wang, X.; Xie, L.; Zhang, X.; Liu, W.; Nie\ss{}ner, M.; and
  Tian, Q. 2022.
\newblock Fast Dynamic Radiance Fields with Time-Aware Neural Voxels.
\newblock In \emph{SIGGRAPH Asia 2022 Conference Papers}.

\bibitem[{Feng et~al.(2021)Feng, Feng, Black, and Bolkart}]{Feng:SIGGRAPH:2021}
Feng, Y.; Feng, H.; Black, M.~J.; and Bolkart, T. 2021.
\newblock Learning an Animatable Detailed {3D} Face Model from In-the-Wild
  Images.
\newblock \emph{ACM Transactions on Graphics (ToG), Proc. SIGGRAPH}, 40(4):
  88:1--88:13.

\bibitem[{Gafni et~al.(2021)Gafni, Thies, Zollhofer, and
  Nie{\ss}ner}]{gafni2021dynamic}
Gafni, G.; Thies, J.; Zollhofer, M.; and Nie{\ss}ner, M. 2021.
\newblock Dynamic neural radiance fields for monocular 4d facial avatar
  reconstruction.
\newblock In \emph{Proceedings of the IEEE/CVF Conference on Computer Vision
  and Pattern Recognition}, 8649--8658.

\bibitem[{Gao et~al.(2022)Gao, Zhong, Xiang, Hong, Guo, and
  Zhang}]{Gao2022nerfblendshape}
Gao, X.; Zhong, C.; Xiang, J.; Hong, Y.; Guo, Y.; and Zhang, J. 2022.
\newblock Reconstructing Personalized Semantic Facial NeRF Models From
  Monocular Video.
\newblock \emph{ACM Transactions on Graphics (Proceedings of SIGGRAPH Asia)},
  41(6).

\bibitem[{Gerig et~al.(2018)Gerig, Morel-Forster, Blumer, Egger, Luthi,
  Sch{\"o}nborn, and Vetter}]{gerig2018morphable}
Gerig, T.; Morel-Forster, A.; Blumer, C.; Egger, B.; Luthi, M.; Sch{\"o}nborn,
  S.; and Vetter, T. 2018.
\newblock Morphable face models-an open framework.
\newblock In \emph{2018 13th IEEE International Conference on Automatic Face \&
  Gesture Recognition (FG 2018)}, 75--82. IEEE.

\bibitem[{Grassal et~al.(2022)Grassal, Prinzler, Leistner, Rother, Nie{\ss}ner,
  and Thies}]{grassal2022neural}
Grassal, P.-W.; Prinzler, M.; Leistner, T.; Rother, C.; Nie{\ss}ner, M.; and
  Thies, J. 2022.
\newblock Neural head avatars from monocular RGB videos.
\newblock In \emph{Proceedings of the IEEE/CVF Conference on Computer Vision
  and Pattern Recognition}, 18653--18664.

\bibitem[{Guo et~al.(2021)Guo, Chen, Liang, Liu, Bao, and Zhang}]{guo2021ad}
Guo, Y.; Chen, K.; Liang, S.; Liu, Y.-J.; Bao, H.; and Zhang, J. 2021.
\newblock AD-NeRF: Audio Driven Neural Radiance Fields for Talking Head
  Synthesis.
\newblock In \emph{Proceedings of the IEEE/CVF International Conference on
  Computer Vision (ICCV)}, 5764--5774.

\bibitem[{Hong et~al.(2022)Hong, Peng, Xiao, Liu, and Zhang}]{hong2022headnerf}
Hong, Y.; Peng, B.; Xiao, H.; Liu, L.; and Zhang, J. 2022.
\newblock HeadNeRF: A Real-Time NeRF-Based Parametric Head Model.
\newblock In \emph{Proceedings of the IEEE/CVF Conference on Computer Vision
  and Pattern Recognition (CVPR)}, 20374--20384.

\bibitem[{Hu et~al.(2017)Hu, Saito, Wei, Nagano, Seo, Fursund, Sadeghi, Sun,
  Chen, and Li}]{hu2017avatar}
Hu, L.; Saito, S.; Wei, L.; Nagano, K.; Seo, J.; Fursund, J.; Sadeghi, I.; Sun,
  C.; Chen, Y.-C.; and Li, H. 2017.
\newblock Avatar Digitization from a Single Image for Real-Time Rendering.
\newblock \emph{ACM Trans. Graph.}, 36(6).

\bibitem[{Ichim, Bouaziz, and Pauly(2015)}]{ichim2015dynamic}
Ichim, A.~E.; Bouaziz, S.; and Pauly, M. 2015.
\newblock Dynamic 3D Avatar Creation from Hand-Held Video Input.
\newblock \emph{ACM Trans. Graph.}, 34(4).

\bibitem[{Kirschstein et~al.(2023)Kirschstein, Qian, Giebenhain, Walter, and
  Nie{\ss}ner}]{kirschstein2023nersemble}
Kirschstein, T.; Qian, S.; Giebenhain, S.; Walter, T.; and Nie{\ss}ner, M.
  2023.
\newblock NeRSemble: Multi-view Radiance Field Reconstruction of Human Heads.
\newblock arXiv:2305.03027.

\bibitem[{Li et~al.(2017)Li, Bolkart, Black, Li, and
  Romero}]{FLAME:SiggraphAsia2017}
Li, T.; Bolkart, T.; Black, M.~J.; Li, H.; and Romero, J. 2017.
\newblock Learning a model of facial shape and expression from {4D} scans.
\newblock \emph{ACM Transactions on Graphics, (Proc. SIGGRAPH Asia)}, 36(6):
  194:1--194:17.

\bibitem[{Li et~al.(2022)Li, Slavcheva, Zollhoefer, Green, Lassner, Kim,
  Schmidt, Lovegrove, Goesele, Newcombe et~al.}]{li2022neural}
Li, T.; Slavcheva, M.; Zollhoefer, M.; Green, S.; Lassner, C.; Kim, C.;
  Schmidt, T.; Lovegrove, S.; Goesele, M.; Newcombe, R.; et~al. 2022.
\newblock Neural 3d video synthesis from multi-view video.
\newblock In \emph{Proceedings of the IEEE/CVF Conference on Computer Vision
  and Pattern Recognition}, 5521--5531.

\bibitem[{Liu et~al.(2022)Liu, Xu, Wu, Zhou, Wu, and Zhou}]{liu2022semantic}
Liu, X.; Xu, Y.; Wu, Q.; Zhou, H.; Wu, W.; and Zhou, B. 2022.
\newblock Semantic-Aware Implicit Neural Audio-Driven Video Portrait
  Generation.
\newblock In \emph{Proceedings of the European Conference on Computer Vision
  (ECCV)}.

\bibitem[{Lombardi et~al.(2018)Lombardi, Saragih, Simon, and
  Sheikh}]{lombardi2018deep}
Lombardi, S.; Saragih, J.; Simon, T.; and Sheikh, Y. 2018.
\newblock Deep Appearance Models for Face Rendering.
\newblock \emph{ACM Trans. Graph.}, 37(4): 68:1--68:13.

\bibitem[{Lombardi et~al.(2019)Lombardi, Simon, Saragih, Schwartz, Lehrmann,
  and Sheikh}]{lombardi2019neural}
Lombardi, S.; Simon, T.; Saragih, J.; Schwartz, G.; Lehrmann, A.; and Sheikh,
  Y. 2019.
\newblock Neural Volumes: Learning Dynamic Renderable Volumes from Images.
\newblock \emph{ACM Trans. Graph.}, 38(4): 65:1--65:14.

\bibitem[{Lombardi et~al.(2021)Lombardi, Simon, Schwartz, Zollhoefer, Sheikh,
  and Saragih}]{lombardi2021mixture}
Lombardi, S.; Simon, T.; Schwartz, G.; Zollhoefer, M.; Sheikh, Y.; and Saragih,
  J. 2021.
\newblock Mixture of Volumetric Primitives for Efficient Neural Rendering.
\newblock \emph{ACM Trans. Graph.}, 40(4).

\bibitem[{Ma et~al.(2021)Ma, Simon, Saragih, Wang, Li, La~Torre, and
  Sheikh}]{ma2021pixel}
Ma, S.; Simon, T.; Saragih, J.; Wang, D.; Li, Y.; La~Torre, F.~D.; and Sheikh,
  Y. 2021.
\newblock Pixel Codec Avatars.
\newblock In \emph{2021 IEEE/CVF Conference on Computer Vision and Pattern
  Recognition (CVPR)}, 64--73.

\bibitem[{Mildenhall et~al.(2020)Mildenhall, Srinivasan, Tancik, Barron,
  Ramamoorthi, and Ng}]{mildenhall2020nerf}
Mildenhall, B.; Srinivasan, P.~P.; Tancik, M.; Barron, J.~T.; Ramamoorthi, R.;
  and Ng, R. 2020.
\newblock NeRF: Representing Scenes as Neural Radiance Fields for View
  Synthesis.
\newblock In \emph{ECCV}.

\bibitem[{Nagano et~al.(2018)Nagano, Seo, Xing, Wei, Li, Saito, Agarwal,
  Fursund, and Li}]{nagano2018pagan}
Nagano, K.; Seo, J.; Xing, J.; Wei, L.; Li, Z.; Saito, S.; Agarwal, A.;
  Fursund, J.; and Li, H. 2018.
\newblock PaGAN: Real-Time Avatars Using Dynamic Textures.
\newblock \emph{ACM Trans. Graph.}, 37(6).

\bibitem[{Park et~al.(2021{\natexlab{a}})Park, Sinha, Barron, Bouaziz, Goldman,
  Seitz, and Martin-Brualla}]{park2021nerfies}
Park, K.; Sinha, U.; Barron, J.~T.; Bouaziz, S.; Goldman, D.~B.; Seitz, S.~M.;
  and Martin-Brualla, R. 2021{\natexlab{a}}.
\newblock Nerfies: Deformable neural radiance fields.
\newblock In \emph{Proceedings of the IEEE/CVF International Conference on
  Computer Vision}, 5865--5874.

\bibitem[{Park et~al.(2021{\natexlab{b}})Park, Sinha, Hedman, Barron, Bouaziz,
  Goldman, Martin-Brualla, and Seitz}]{park2021hypernerf}
Park, K.; Sinha, U.; Hedman, P.; Barron, J.~T.; Bouaziz, S.; Goldman, D.~B.;
  Martin-Brualla, R.; and Seitz, S.~M. 2021{\natexlab{b}}.
\newblock Hypernerf: A higher-dimensional representation for topologically
  varying neural radiance fields.
\newblock \emph{arXiv preprint arXiv:2106.13228}.

\bibitem[{Pumarola et~al.(2021)Pumarola, Corona, Pons-Moll, and
  Moreno-Noguer}]{pumarola2021d}
Pumarola, A.; Corona, E.; Pons-Moll, G.; and Moreno-Noguer, F. 2021.
\newblock D-nerf: Neural radiance fields for dynamic scenes.
\newblock In \emph{Proceedings of the IEEE/CVF Conference on Computer Vision
  and Pattern Recognition}, 10318--10327.

\bibitem[{{Sara Fridovich-Keil and Giacomo Meanti} et~al.(2023){Sara
  Fridovich-Keil and Giacomo Meanti}, Warburg, Recht, and
  Kanazawa}]{kplanes_2023}
{Sara Fridovich-Keil and Giacomo Meanti}; Warburg, F.~R.; Recht, B.; and
  Kanazawa, A. 2023.
\newblock K-Planes: Explicit Radiance Fields in Space, Time, and Appearance.
\newblock In \emph{CVPR}.

\bibitem[{Sun et~al.(2022)Sun, Wang, Shi, Wang, Wang, and Liu}]{sun2022ide}
Sun, J.; Wang, X.; Shi, Y.; Wang, L.; Wang, J.; and Liu, Y. 2022.
\newblock IDE-3D: Interactive Disentangled Editing for High-Resolution 3D-aware
  Portrait Synthesis.
\newblock \emph{ACM Transactions on Graphics (TOG)}, 41(6): 1--10.

\bibitem[{Sun et~al.(2023)Sun, Wang, Wang, Li, Zhang, Zhang, and
  Liu}]{sun2023next3d}
Sun, J.; Wang, X.; Wang, L.; Li, X.; Zhang, Y.; Zhang, H.; and Liu, Y. 2023.
\newblock Next3D: Generative Neural Texture Rasterization for 3D-Aware Head
  Avatars.
\newblock In \emph{Proceedings of the IEEE/CVF Conference on Computer Vision
  and Pattern Recognition (CVPR)}.

\bibitem[{Wang et~al.(2022)Wang, Chandran, Zoss, Bradley, and
  Gotardo}]{wang2022morf}
Wang, D.; Chandran, P.; Zoss, G.; Bradley, D.; and Gotardo, P. 2022.
\newblock MoRF: Morphable Radiance Fields for Multiview Neural Head Modeling.
\newblock In \emph{ACM SIGGRAPH 2022 Conference Proceedings}, SIGGRAPH '22. New
  York, NY, USA: Association for Computing Machinery.
\newblock ISBN 9781450393379.

\bibitem[{Wang et~al.(2021)Wang, Liu, Liu, Theobalt, Komura, and
  Wang}]{wang2021neus}
Wang, P.; Liu, L.; Liu, Y.; Theobalt, C.; Komura, T.; and Wang, W. 2021.
\newblock NeuS: Learning Neural Implicit Surfaces by Volume Rendering for
  Multi-view Reconstruction.
\newblock \emph{NeurIPS}.

\bibitem[{Xu et~al.(2023)Xu, Wang, Zhao, Zhang, and Liu}]{xu2023avatarmav}
Xu, Y.; Wang, L.; Zhao, X.; Zhang, H.; and Liu, Y. 2023.
\newblock AvatarMAV: Fast 3D Head Avatar Reconstruction Using Motion-Aware
  Neural Voxels.
\newblock In \emph{ACM SIGGRAPH 2023 Conference Proceedings}.

\bibitem[{Zhang et~al.(2018)Zhang, Isola, Efros, Shechtman, and
  Wang}]{zhang2018unreasonable}
Zhang, R.; Isola, P.; Efros, A.~A.; Shechtman, E.; and Wang, O. 2018.
\newblock The unreasonable effectiveness of deep features as a perceptual
  metric.
\newblock In \emph{Proceedings of the IEEE conference on computer vision and
  pattern recognition}, 586--595.

\bibitem[{Zheng et~al.(2022)Zheng, Abrevaya, B{\"u}hler, Chen, Black, and
  Hilliges}]{zheng2022avatar}
Zheng, Y.; Abrevaya, V.~F.; B{\"u}hler, M.~C.; Chen, X.; Black, M.~J.; and
  Hilliges, O. 2022.
\newblock Im avatar: Implicit morphable head avatars from videos.
\newblock In \emph{Proceedings of the IEEE/CVF Conference on Computer Vision
  and Pattern Recognition}, 13545--13555.

\bibitem[{Zheng et~al.(2023)Zheng, Yifan, Wetzstein, Black, and
  Hilliges}]{Zheng2023pointavatar}
Zheng, Y.; Yifan, W.; Wetzstein, G.; Black, M.~J.; and Hilliges, O. 2023.
\newblock PointAvatar: Deformable Point-based Head Avatars from Videos.
\newblock In \emph{Proceedings of the IEEE/CVF Conference on Computer Vision
  and Pattern Recognition (CVPR)}.

\bibitem[{Zhuang et~al.(2022)Zhuang, Zhu, Sun, and Cao}]{zhuang2022mofanerf}
Zhuang, Y.; Zhu, H.; Sun, X.; and Cao, X. 2022.
\newblock MoFaNeRF: Morphable Facial Neural Radiance Field.
\newblock In \emph{Proceedings of the European Conference on Computer Vision
  (ECCV)}.

\bibitem[{Zielonka, Bolkart, and Thies(2022)}]{MICA:ECCV2022}
Zielonka, W.; Bolkart, T.; and Thies, J. 2022.
\newblock Towards Metrical Reconstruction of Human Faces.

\bibitem[{Zielonka, Bolkart, and Thies(2023)}]{INSTA:CVPR2023}
Zielonka, W.; Bolkart, T.; and Thies, J. 2023.
\newblock Instant Volumetric Head Avatars.

\bibitem[{zllrunning(2019)}]{faceparse}
zllrunning. 2019.
\newblock face-parsing.PyTorch.

\end{thebibliography}

\end{document}